\documentclass{new_tlp}
\usepackage{mathptmx}
\usepackage{times}
\usepackage{helvet}
\usepackage{courier}
\usepackage{amsmath}
\usepackage{amssymb}
\usepackage{Tabbing}
\usepackage{algorithm}
\usepackage{natbib}
\usepackage{graphicx}
\usepackage{epstopdf}
\usepackage{verbatim}
\usepackage{url}
\usepackage{graphicx}
\usepackage{algorithmicx}
\usepackage{algpseudocode}

  \title[Theory and Practice of Logic Programming ]
        {Efficient Computation of the Well-Founded Semantics over Big Data}

  \author[Ilias Tachmazidis, Grigoris Antoniou and Wolfgang Faber]
         {Ilias Tachmazidis, Grigoris Antoniou and Wolfgang Faber\\
         University of Huddersfield, UK\\
         \email{\{ilias.tachmazidis,g.antoniou,w.faber\}@hud.ac.uk}}

\jdate{March 2003}
\pubyear{2003}
\pagerange{\pageref{firstpage}--\pageref{lastpage}}
\doi{S1471068401001193}

\newtheorem{definition}{Definition}[section]
\newtheorem{lemma}{Lemma}[section]
\newtheorem{theorem}{Theorem}[section]

\begin{document}

\label{firstpage}

\maketitle

  \begin{abstract}
    Data originating from the Web, sensor readings and social media result in increasingly huge
    datasets. The so called Big Data comes with new scientific and technological challenges
    while creating new opportunities, hence the increasing interest in academia and industry.
    Traditionally, logic programming has focused on complex knowledge structures/programs, so the
    question arises whether and how it can work in the face of Big Data. In this paper, we examine
    how the well-founded semantics can process huge amounts of data through mass parallelization.
    More specifically, we propose and evaluate a parallel approach using the MapReduce framework.
    Our experimental results indicate that our approach is scalable and that well-founded semantics
    can be applied to billions of facts. To the best of our knowledge, this is the first work that
    addresses large scale nonmonotonic reasoning without the restriction of stratification for predicates
    of arbitrary arity.
  \end{abstract}



\section{Introduction}

Huge amounts of data are being generated at an increasing pace by sensor networks, government authorities
and social media. Such data is heterogeneous, and often needs to be combined with other information,
including database and web data, in order to become more useful. This {\em big data} challenge is at the core of many
contemporary scientific, technological and business developments.

The question arises whether the reasoning community, as found in the areas of knowledge representation,
rule systems, logic programming and semantic web, can connect to the big data wave. On the one hand,
there is a clear application scope, e.g. deriving higher-level knowledge, assisting decision support and data
cleaning. But on the other hand, there are significant challenges arising from the area's
traditional focus on rich knowledge structures instead of large amounts of data, and its reliance on
in-memory methods. The best approach for enabling reasoning with big data is parallelization, as
established e.g. by the LarKC project~(\cite{DBLP:conf/semco/FenselHABCVFHKLSTWWZ08}).

As discussed in~(\cite{DBLP:conf/semco/FenselHABCVFHKLSTWWZ08}), reasoning on the large scale can be achieved
through parallelization by distributing the computation among nodes. There are mainly two proposed approaches
in the literature, namely rule partitioning and data partitioning~(\cite{DBLP:conf/icpp/SomaP08}).

In the case of rule partitioning, the computation of each rule is assigned to a node in the cluster.
Thus, the workload for each rule (and node) depends on the structure and the size of the given rule
set, which could possibly prevent balanced work distribution and high scalability.
On the other hand, for the case of data partitioning, data is divided in chunks with each chunk assigned to a
node, allowing more balanced distribution of the computation among nodes.

Parallel reasoning, based on data partitioning, has been studied extensively. 
In particular, MARVIN~(\cite{marvin}), follows the \emph{divide-conquer-swap} strategy in which triples are
being swapped between nodes in the cluster in order to achieve balanced workload. MARVIN implements
the SpeedDate method, presented in~(\cite{DBLP:conf/www/KotoulasOH10}), where authors pointed out and addressed the scalability
challenge posed by the highly uneven distribution of Semantic Web data.

WebPIE~(\cite{j.websem222}) implements forward reasoning under RDFS and OWL \emph{ter Horst} semantics
over the MapReduce framework~(\cite{Dean:2004:MSD:1251254.1251264}) scaling up to 100 billion triples. In~(\cite{DBLP:conf/esws/GoodmanJMAAH11}) authors
present RDFS inference scaling up to 512 processors with the ability to process, entirely in-memory, 20 billion triples.

FuzzyPD~(\cite{Liu:2011:LSF:2063016.2063043,DBLP:journals/cim/LiuQWY12}) is a MapReduce based prototype
system allowing fuzzy reasoning in OWL $pD^{*}$ with scalability of up to 128 process units and over 1 billion
triples.
Description logic in the form of $\mathcal{EL+}$ have been studied in~(\cite{DBLP:conf/dlog/MutharajuMH10}).
The authors parallelize an existing algorithm for $\mathcal{EL+}$ classification by converting it into MapReduce
algorithms, while experimental evaluation was deferred to future work.

(\cite{DBLP:conf/kr/TachmazidisAFK12}) deals with defeasible logic for unary predicates scaling up to billions of
facts, while authors extend their approach in~(\cite{DBLP:conf/ecai/TachmazidisAFKM12}) for predicates of arbitrary arity,
under the assumption of stratification, scaling up to millions of facts. Finally, the computation of stratified semantics
of logic programming that can be applied to billions of facts is reported in~(\cite{DBLP:conf/ruleml/TachmazidisA13}).

In this paper, we propose a parallel approach for the well-founded semantics computation using the MapReduce framework.
Specifically, we adapt and incorporate the computation of joins and anti-joins, initially described in~(\cite{DBLP:conf/ruleml/TachmazidisA13}).
The crucial difference is that in this paper recursion through negation is allowed, meaning that the well-founded model can contain undefined atoms.
A challenge in this respect is that materializing the Herbrand base is impractical in the context of big data. 
 To overcome this scalability
barrier we require programs to be safe 
and apply a reasoning procedure that allows closure calculation based on the consecutive computation of true
and unknown literals, requiring significantly less information. 
Experimental results highlight the advantages of the applied optimizations, while showing that our approach can scale up
to 1 billion facts even on a modest computational setup.

The rest of the paper is organized as follows. Section~\ref{sec:Preliminaries}
introduces briefly the MapReduce framework, the well-founded semantics and the alternating
fixpoint procedure. Join and anti-join operations for the well-founded semantics are described in
Section~\ref{sec:Computing_TPJI}. Section~\ref{sec:Computing the Well-Founded Semantics}
provides a parallel implementation over the MapReduce framework, while
experimental results are presented in Section~\ref{sec:Experimental_results}. We conclude and
discuss future directions in Section~\ref{sec:Conclusion_and_Future_Work}.

\section{Preliminaries}\label{sec:Preliminaries}

\subsection{MapReduce Framework}\label{sec:MapReduce_Framework}
MapReduce is a framework for parallel processing over huge datasets~(\cite{Dean:2004:MSD:1251254.1251264}).
Processing is carried out in a map and a reduce phase. For each phase, a set of user-defined map
and reduce functions are run in parallel. The former performs a user-defined operation over an arbitrary part
of the input and partitions the data, while the latter performs a user-defined operation on each partition.

MapReduce is designed to operate over key/value pairs. Specifically, each $Map$ function receives a key/value
pair and emits a set of key/value pairs. Subsequently, all key/value pairs produced during the map phase are
grouped by their key and passed to reduce phase. During the reduce phase, a $Reduce$ function is called for
each unique key, processing the corresponding set of values.

Let us illustrate the \emph{wordcount} example. In this example, we take as input a large number of documents
and calculate the frequency of each word. The pseudo-code for the $Map$ and $Reduce$ functions is provided
in~\ref{sec:MapReduce_algorithms}.


Consider the following documents as input:
\begin{center}
	\begin{tabular}{ l }
        Doc1: ``Hello world.'' \\
        Doc2: ``Hello MapReduce.''
	\end{tabular}
\end{center}

During map phase, each map operation gets as input a line of a document.
$Map$ function extracts words from each line and emits pairs of the form
$<$w, ``1''$>$ meaning that word \emph{w} occurred once (``1''), namely
the following pairs:
\begin{center}
	\begin{tabular}{ l l l l }
        $<$Hello, 1$>$ & $<$world, 1$>$ & $<$Hello, 1$>$ & $<$MapReduce, 1$>$ 
	\end{tabular}
\end{center}

MapReduce framework will perform grouping/sorting resulting in the
following intermediate pairs:
\begin{center}
	\begin{tabular}{ c c c }
        $<$Hello, $<$1,1$>$$>$ & $<$world, 1$>$ & $<$MapReduce, 1$>$
	\end{tabular}
\end{center}

During the reduce phase, the $Reduce$ function sums up all
occurrence values for each word emitting a pair containing the word
and the frequency of the word. Thus, the reducer with key:
\begin{center}
	\begin{tabular}{ l }
        \emph{Hello} will emit \emph{$<$Hello, 2$>$} \\
        \emph{world} will emit $<$world, 1$>$ \\
        \emph{MapReduce} will emit $<$MapReduce, 1$>$
	\end{tabular}
\end{center}

\subsection{Well-Founded Semantics}\label{sec:Well-Founded_Semantics}
In this section we provide the definition of the \emph{well-founded semantics} (WFS)
as it was defined in~(\cite{DBLP:journals/jacm/GelderRS91}).

\begin{definition}\label{def:general_logic_program}
(\cite{DBLP:journals/jacm/GelderRS91}) A \emph{general logic program} is a finite
set of \emph{general rules}, which may have both positive and negative
subgoals. A general rule is written with its \emph{head}, or conclusion
on the left, and its subgoal (body), if any to the right of the symbol
``$\leftarrow$'', which may be read ``if''. For example,
\begin{center}
	\begin{tabular}{ l }
        p(X) $\leftarrow$ a(X), {\bf not} b(X).
	\end{tabular}
\end{center}
is a rule in which \emph{p(X)} is the head, \emph{a(X)} is a
\emph{positive subgoal}, and \emph{b(X)} is a \emph{negative subgoal}.
This rule may be read as ``\emph{p(X)} if \emph{a(X)} and not
\emph{b(X)}''. A \emph{Horn rule} is one with no negative subgoals,
and a \emph{Horn logic program} is one with only Horn rules.
\proofbox \end{definition}

We use the following conventions. A logical variable starts with a
capital letter while a constant or a predicate starts with a lowercase
letter. Note that functions are not allowed. A predicate of arbitrary
arity will be referred as a \emph{literal}. Constants, variables and
literals are \emph{terms}. A \emph{ground term} is a term with no
variables. The \emph{Herbrand universe} is the set of constants in a
given program. The \emph{Herbrand base} is the set of ground terms
that are produced by the substitution of variables with constants in
the Herbrand universe. In this paper, we will refer to Horn rules also
as \emph{definite} rules, likewise Horn programs will also be referred
to as \emph{definite} programs.

%

\begin{definition}\label{def:interpretation_I}
(\cite{DBLP:journals/jacm/GelderRS91})
Given a program {\bf P}, a \emph{partial interpretation I} is a
consistent set of literals whose atoms are in the Herbrand base of
{\bf P}. A \emph{total interpretation} is a partial interpretation
that contains every atom of the Hebrand base or its negation. We say
a ground (variable-free) literal is \emph{true in I} when it is in
\emph{I} and say it is \emph{false in I} when its complement is in
\emph{I}. Similarly, we say a conjunction of ground literals is
\emph{true} in \emph{I} if all of the literals are true in \emph{I},
and is \emph{false} in \emph{I} if any of its literals is false in
\emph{I}.
\proofbox \end{definition}

\begin{definition}\label{def:unfounded sets}
(\cite{DBLP:journals/jacm/GelderRS91})
Let a program {\bf P}, its associated Herbrand base \emph{H} and a
partial interpretation \emph{I} be given. We say \emph{A} $\subseteq$
\emph{H} is an \emph{unfounded set} (of {\bf P}) with respect to \emph{I}
if each atom \emph{p} $\in$ \emph{A} satisfies the following
condition: For each instantiated rule $R$ of {\bf P} whose head
is \emph{p}, (at least) one of the following holds:
\begin{enumerate}
  \item Some (positive or negative) subgoal of the body is
    false in \emph{I}.
  \item Some positive subgoal of the body occurs in \emph{A}.
\end{enumerate}
A literal that makes (1) or (2) above true is called a \emph{witness of
unusability} for rule $R$ (with respect to \emph{I}).
\proofbox \end{definition}

\begin{theorem}\label{WFS_complexity}
(\cite{DBLP:journals/jacm/GelderRS91})
The data complexity of the well-founded semantics for function-free programs is polynomial time.
\proofbox \end{theorem}

In this paper, we require each rule to be safe, that is, each variable
in a rule must occur (also) in a positive subgoal. Safe programs
consist of safe rules only. This safety criterion is an adaptation of
range restriction (\cite{DBLP:journals/acta/Nicolas82}), which
guarantees the important concept of domain independence, originally
studied in deductive databases (see for example
(\cite{DBLP:books/aw/AbiteboulHV95})).
Apart from this semantic property, the safety condition
implicitly also enforces a certain locality of computation, which is
important for our proposed method, as we shall discuss in
Section~\ref{sec:Final_remarks}.

\subsection{Alternating Fixpoint Procedure}\label{sec:Alternating_Fixpoint_Procedure}
In this section, we provide the definition of the alternating
fixpoint procedure as it was defined in~(\cite{bradixfrezuk99}).

\begin{definition}\label{Positive_Negative_Literals}
(\cite{bradixfrezuk99})
For a set \emph{S} of literals we define the following sets:
\begin{center}
	\begin{tabular}{ l l l}
        pos(S) & \emph{:=} & \{\emph{A} $\in$ S $|$ \emph{A} is a positive literal \},\\
        neg(S) & \emph{:=} & \{\emph{A} $|$ {\bf not} \emph{A} $\in$ \emph{S}\}. \proofbox
	\end{tabular}
\end{center}
\end{definition}

\begin{definition}\label{Extended_Immediate_Consequence_Operator}
(\cite{bradixfrezuk99})
(\emph{Extended Immediate Consequence Operator})\\
Let \emph{P} be a normal logic program. Let \emph{I} and \emph{J} be sets of ground
atoms. The set \emph{T$_{P,J}$(I)} of \emph{immediate consequences} of \emph{I} w.r.t. \emph{P}
and \emph{J} is defined as follows:
\begin{center}
	\begin{tabular}{ l }
        \emph{T$_{P,J}$(I)} \emph{:=} \{\emph{A} $|$ there is \emph{A} $\leftarrow$ $\mathcal{B}$ $\in$ ground(\emph{P}) with
        pos($\mathcal{B}$) $\subseteq$ \emph{I} and neg($\mathcal{B}$) $\cap$ \emph{J} = $\emptyset$\}.
	\end{tabular}
\end{center}
If \emph{P} is definite, the set \emph{J} is not needed and we obtain the standard immediate
consequence operator T$_{P}$ by \emph{T$_{P}$(I)} $=$ \emph{T$_{P,\emptyset}$(I)}.
\proofbox \end{definition}

For an operator $T$ we define $T$ $\uparrow$ $0$ $:=$ $\emptyset$ and $T$ $\uparrow$ $i$ $:=$
$T(T \uparrow i-1)$, for $i > 0$. lfp($T$) denotes the least fixpoint of $T$, i.e. the smallest
set $S$ such that $T(S) = S$.

\begin{definition}\label{Alternating_Fixpoint_Procedure}
(\cite{bradixfrezuk99})
(\emph{Alternating Fixpoint Procedure})\\
Let \emph{P} be a normal logic program. Let \emph{P$^{+}$} denote the subprogram consisting of
the definite rules of \emph{P}. Then the sequence (K$_{i}$,U$_{i}$)$_{i\geq0}$ with set \emph{K$_{i}$}
of true (known) facts and \emph{U$_{i}$} of possible (unknown) facts is defined by:\\
\begin{center}
	\begin{tabular}{ l l l }
        ~ & K$_{0}$ $:=$ lfp(T$_{P^{+}}$) & U$_{0}$ $:=$ lfp(T$_{P,K_{0}}$) \\
        $i>0:$ & K$_{i}$ $:=$ lfp(T$_{P,U_{i-1}}$) & U$_{i}$ $:=$ lfp(T$_{P,K_{i}}$)
	\end{tabular}
\end{center}
The computation terminates when the sequence becomes stationary, i.e., when a
fixpoint is reached in the sense that (K$_{i}$,U$_{i}$) = (K$_{i+1}$,U$_{i+1}$).
This computation schema is called the \emph{Alternating Fixpoint Procedure} (AFP).
\proofbox \end{definition}

We rely on the definition of the \emph{well-founded partial model W$^{*}_{p}$} of P as given
in~(\cite{DBLP:journals/jacm/GelderRS91}).

\begin{theorem}\label{Correctness_of_AFP}
(\cite{bradixfrezuk99})
(Correctness of \emph{AFP})\\
Let the sequence (K$_{i}$,U$_{i}$)$_{i\geq0}$ be defined as above. Then there is a
$j\geq0$ such that (K$_{j}$,U$_{j}$)= (K$_{j+1}$,U$_{j+1}$). The well-founded
model \emph{W$^{*}_{p}$}  of \emph{P} can be directly derived from the fixpoint
(K$_{j}$,U$_{j}$), i.e.,
\begin{center}
	\begin{tabular}{ l l }
        W$^{*}_{p}$ = \{L $|$ & L is a positive ground literal and \emph{L $\in$ K$_{j}$} or \\
        ~ & L is a negative ground literal \emph{{\bf not} A} and \emph{A $\in$ BASE(P) $-$ U$_{j}$}\},
	\end{tabular}
\end{center}
where BASE(P) is the Herbrand base of program P.
\proofbox \end{theorem}

\begin{lemma}\label{Monotonicity}
(\cite{bradixfrezuk99})
(\emph{Monotonicity})\\
Let the sequence (K$_{i}$,U$_{i}$)$_{i\geq0}$ be defined as above. Then the following
holds for $i\geq0:$\\
K$_{i}$ $\subseteq$ K$_{i+1}$, U$_{i}$ $\supseteq$ U$_{i+1}$, K$_{i}$ $\subseteq$ U$_{i}$.
\proofbox \end{lemma}


\section{Computing \emph{T$_{P,J}$(I)}}\label{sec:Computing_TPJI}

Consider the following program:
\begin{center}
	\begin{tabular}{ l }
        p(X,Y) $\leftarrow$ a(X,Z), b(Z,Y), {\bf not} c(X,Z), {\bf not} d(Z,Y).
	\end{tabular}
\end{center}
Here \emph{p(X,Y)} is our \emph{final goal}, \emph{a(X,Z)} and \emph{b(Z,Y)}
are positive subgoals, while \emph{c(X,Z)} and \emph{d(Z,Y)}
are negative subgoals. In order to compute our final goal \emph{p(X,Y)} we need
to ensure that \{\emph{a(X,Z)}, \emph{b(Z,Y)}\} $\subseteq$ \emph{I} and \{\emph{c(X,Z)},
\emph{d(Z,Y)}\} $\cap$ \emph{J} = $\emptyset$ (see Definition~\ref{Extended_Immediate_Consequence_Operator}),
namely both \emph{a(X,Z)} and \emph{b(Z,Y)}
are in \emph{I} while none of \emph{c(X,Z)} and \emph{d(Z,Y)} is found in \emph{J}.

As positive subgoals depend on \emph{I} we can group them into a \emph{positive goal}.
A positive goal consists of a new predicate (say \emph{ab}) that contains as arguments
the union of two sets: (a) all the arguments of the final goal (X,Y) and (b) all the
common arguments between positive and negative subgoals (X,Z,Y), namely we need to
compute \emph{ab(X,Z,Y)}. The final goal (\emph{p(X,Y)}) consists of all values of the
positive goal (\emph{ab(X,Z,Y)}) that do not match any of the negative subgoals (\emph{c(X,Z)}
and \emph{d(Z,Y)}) on their common arguments (X,Z
and Z,Y respectively).

\subsection{Positive goal calculation}\label{sec:Positive_goal_calculation}
Consider the following program:
\begin{center}
	\begin{tabular}{ l }
        p(X,Y) $\leftarrow$ a(X,Z), b(Z,Y), {\bf not} c(X,Z), {\bf not} d(Z,Y).
	\end{tabular}
\end{center}
where
\begin{center}
	\begin{tabular}{ l }
        \emph{I} = \{a(1,2), a(1,3), b(2,4), b(3,5)\}  \\
        \emph{J} = \{c(1,2), d(2,3)\}
	\end{tabular}
\end{center}

A single join~(\cite{Cluet94classificationand}), calculating the positive goal \emph{ab(X,Z,Y)}, can be performed
as described below. 
The pseudo-code for the $Map$ and $Reduce$ functions is provided in~\ref{sec:MapReduce_algorithms}.
Note that we use only literals from \emph{I}.


The $Map$ function will emit pairs of the form $<$Z,(a,X)$>$ for predicate
\emph{a} and $<$Z,(b,Y)$>$ for predicate \emph{b}, namely the following pairs:
\begin{center}
	\begin{tabular}{ c c c c}
        $<$2, (a,1)$>$ & $<$3, (a,1)$>$ & $<$2, (b,4)$>$ & $<$3, (b,5)$>$
	\end{tabular}
\end{center}

MapReduce framework will perform grouping/sorting resulting in the
following intermediate pairs:
\begin{center}
	\begin{tabular}{ c c }
        $<$2, $<$(a,1), (b,4)$>$$>$ & $<$3, $<$(a,1), (b,5)$>$$>$
	\end{tabular}
\end{center}

During the reduce phase we match predicates \emph{a} and \emph{b} on their
common argument (which is the \emph{key}) and use the values to
emit positive goals. Thus, the reducer with key:
\begin{center}
	\begin{tabular}{ c }
        \emph{2} will emit \emph{ab(1,2,4)} \\
        \emph{3} will emit \emph{ab(1,3,5)}
	\end{tabular}
\end{center}

Note that we need to filter out possibly occurring duplicates as soon
as possible because they will produce unnecessary duplicates as well,
affecting the overall performance. The pseudo-code and a brief description
of duplicate elimination are provided in~\ref{sec:MapReduce_algorithms}.



For rules with more than one join between positive subgoals we need to
apply multi-joins (multi-way join).

Consider the following program:
\begin{center}
	\begin{tabular}{ l }
        q(X,Y) $\leftarrow$ a(X,Z), b(Z,W), c(W,Y), {\bf not} d(X,W).
	\end{tabular}
\end{center}
We can compute the positive goal (\emph{abc(X,W,Y)}) by applying our approach
for single join twice. First, we need to join \emph{a(X,Z)} and \emph{b(Z,W)}
on \emph{Z}, producing a temporary literal (say \emph{ab(X,W)}), and then join
\emph{ab(X,W)} and \emph{c(W,Y)} on \emph{W} producing the positive goal
(\emph{abc(X,W,Y)}). Once \emph{abc(X,W,Y)} is calculated, we proceed with
calculating the final goal \emph{q(X,Y)} by retaining all the values of
\emph{abc(X,W,Y)} that do not match \emph{d(X,W)} on their common
arguments (X,W).

For details on single and multi-way join, readers are referred to literature.
More specifically, multi-way join has been described and
optimized in~(\cite{DBLP:conf/edbt/AfratiU10}). In order to achieve an efficient
implementation, optimizations in~(\cite{DBLP:conf/edbt/AfratiU10}) should be taken
into consideration.

\subsection{Final goal calculation}\label{sec:Final_goal_calculation}
Consider the program mentioned at the beginning of Section~\ref{sec:Positive_goal_calculation}.
By calculating the positive goal \emph{ab(X,Z,Y)} we obtain the following knowledge:
\begin{center}
	\begin{tabular}{ c c c }
        \emph{ab(1,2,4)} & ~ & \emph{ab(1,3,5)}
	\end{tabular}
\end{center}

In order to calculate the final goal (\emph{p(X,Y)}) we need to perform an
anti-join~(\cite{Cluet94classificationand}) between \emph{ab(X,Z,Y)} and each negative subgoal (\emph{c(X,Z)}
and \emph{d(Z,Y)}). Note that to perform an anti-join we use only the
previously calculated positive goal (\emph{ab(X,Z,Y)}) and literals from \emph{J}.

We start by performing an anti-join between \emph{ab(X,Z,Y)} and \emph{c(X,Z)}
on their common arguments (X,Z), creating a new literal (say \emph{abc(X,Z,Y)}), which
contains all the results from \emph{ab(X,Z,Y)} that are not found in \emph{c(X,Z)}, as
described below. 
The pseudo-code for the $Map$ and $Reduce$ functions is provided in~\ref{sec:MapReduce_algorithms}.


The $Map$ function will emit pairs of the form $<$(X,Z),(ab,Y)$>$ for predicate
\emph{ab} and $<$(X,Z),c$>$ for predicate \emph{c} (while predicate \emph{d}
will be taken into consideration during the next anti-join), namely the following
pairs:
\begin{center}
	\begin{tabular}{ c c c }
        $<$(1,2), (ab,4)$>$ & $<$(1,3), (ab,5)$>$ & $<$(1,2), c$>$
 	\end{tabular}
\end{center}

MapReduce framework will perform grouping/sorting resulting in the
following intermediate pairs:
\begin{center}
	\begin{tabular}{ c c }
        $<$(1,2), $<$(ab,4), (c)$>$$>$ & $<$(1,3), (ab,5)$>$
	\end{tabular}
\end{center}

During the reduce phase we output values of the predicate \emph{ab}
only if it is not matched by predicate \emph{c} on their common arguments
(which are contained in the \emph{key}) and emit
\emph{abc(X,Z,Y)}. Thus, the reducer with key:
\begin{center}
	\begin{tabular}{ c }
        \emph{(1,2)} will have no output \\
        \emph{(1,3)} will emit \emph{abc(1,3,5)}
	\end{tabular}
\end{center}

In order to calculate the final goal (\emph{p(X,Y)}), we need to perform an additional
anti-join between \emph{abc(X,Z,Y)} and \emph{d(Z,Y)} on their common arguments (Z,Y).
Here, \emph{abc(1,3,5)} and \emph{d(2,3)} do not match on their common arguments (Z,Y)
as (3,5) $\neq$ (2,3). Thus, our calculated final goal is \emph{p(1,5)}.

\section{Computing the Well-Founded Semantics}\label{sec:Computing the Well-Founded Semantics}
In this section we describe 
an optimized implementation for the
calculation of the well-founded semantics. A naive implementation is considered as
one following Definition~\ref{Alternating_Fixpoint_Procedure} while ignoring the monotonicity
properties of the well-founded semantics (see Lemma~\ref{Monotonicity}).

\subsection{Optimized implementation}\label{sec:Optimized_implementation}

A naive implementation would introduce unnecessary overhead to
the overall computation since it comes with the overhead of reasoning over and
storage of overlapping sets of knowledge. A more refined version of both WFS fixpoint and least
fixpoint of \emph{T$_{P,J}$(I)} is defined in Algorithm~\ref{Optimized_WFS_fixpoint}
and Algorithm~\ref{Optimized_least fixpoint} respectively.

\begin{algorithm}[htbp]
\begin{algorithmic}[1]
\Statex opt\_WFS\_fixpoint(\emph{P}): \Comment input: program \emph{P}
\State \emph{K$_{0}$} = opt\_lfp(\emph{P+}, \emph{$\emptyset$}, \emph{$\emptyset$}); \Comment output: set of literals \emph{K$_{i-1}$}, \emph{U$_{i-1}$}
\State \emph{i} = 0;
\Repeat
\State \emph{U$_{i}$} = \emph{K$_{i}$} $\cup$ opt\_lfp(\emph{P}, \emph{K$_{i}$}, \emph{K$_{i}$});
\State \emph{i}++; \Comment next ``inference step''
\State \emph{K$_{i}$} = \emph{K$_{i-1}$} $\cup$ opt\_lfp(\emph{P}, K$_{i-1}$, \emph{U$_{i-1}$});
\Until{\emph{K$_{i-1}$}.size() == \emph{K$_{i}$}.size()}
\State \Return \emph{K$_{i-1}$}, \emph{U$_{i-1}$};
\end{algorithmic}
\caption{Optimized WFS fixpoint}
\label{Optimized_WFS_fixpoint}
\end{algorithm}

\begin{algorithm}[htbp]
\begin{algorithmic}[1]
\Statex opt\_lfp(\emph{P}, \emph{I}, \emph{J}): \Comment precondition: \emph{I} $\subseteq$ lfp(\emph{T$_{P,J}$($\emptyset$)})
\State \emph{S} = $\emptyset$; \Comment input: program \emph{P}, set of literals \emph{I} and \emph{J}
\State \emph{new} = $\emptyset$; \Comment output: set of literals \emph{S} (lfp(\emph{T$_{P,J}$(I)} - \emph{I})
\Repeat
\State \emph{S} = \emph{S} $\cup$ \emph{new};
\State \emph{new} = T(\emph{P}, (\emph{I} $\cup$ \emph{S}), \emph{J});
\State \emph{new} = \emph{new} - (\emph{I} $\cup$ \emph{S});
\Until{\emph{new} == $\emptyset$}
\State \Return \emph{S};
\end{algorithmic}
\caption{Optimized least fixpoint of \emph{T$_{P,J}$(I)}}
\label{Optimized_least fixpoint}
\end{algorithm}



Our first optimization is the changed calculation of the least fixpoint of
\emph{T$_{P,J}$(I)} (\emph{opt\_lfp}), which is depicted in Algorithm~\ref{Optimized_least fixpoint}.
Instead of calculating the least fixpoint starting from \emph{I} $=$ $\emptyset$,
for a given program \emph{P} and a set of literals \emph{J}, 
we allow the calculation to start
from a given \emph{I}, provided that \emph{I} $\subseteq$ \emph{lfp(T$_{P,J}$($\emptyset$))},
and return only the newly inferred literals (\emph{S})
that led us to the least fixpoint. Thus, the actual set of literals that the
least fixpoint of \emph{T$_{P,J}$(I)} consists of is \emph{I} $\cup$ \emph{S}.
In order to reassure correctness we need to take into consideration both
\emph{I} and \emph{S} while calculating the least fixpoint, namely new literals
are inferred by calculating \emph{T$_{P,J}$(I $\cup$ S)}. However, 
we use a temporary set of inferred literals (\emph{new}) in order to eliminate duplicates
(\emph{new}~$=$~\emph{new}~$-$~(\emph{I}~$\cup$~\emph{S})) prior to adding newly
inferred literals to the set \emph{S} (\emph{S}~$=$~\emph{S}~$\cup$ \emph{new}). Note
that the set of literals \emph{I} remains unchanged when the optimized least fixpoint
is calculated.

The optimized version of the least fixpoint is used, in Algorithm~\ref{Optimized_WFS_fixpoint},
for the computation of each set of literals \emph{K} and \emph{U}. \emph{K$_{0}$} is a special case where
we start from \emph{I} $=$ $\emptyset$ and \emph{J} $=$ $\emptyset$, and thus, unable to
fully utilize the advantages of the optimized least fixpoint.

The proposed optimizations are mainly based on the monotonicity of the well-founded semantics
as given in Lemma~\ref{Monotonicity}. Note that in this section, the indices of the sets \emph{K} and
\emph{U} found in Lemma~\ref{Monotonicity} are adjusted to the indices used in
Algorithm~\ref{Optimized_WFS_fixpoint} in order to facilitate our discussion.

Since \emph{K$_{i}$} $\subseteq$ \emph{U$_{i}$}, for $i\geq0$ (see Lemma~\ref{Monotonicity}), the
computation of \emph{U$_{i}$} can start from \emph{K$_{i}$}, namely \emph{I} $=$ \emph{K$_{i}$}.
Thus, instead of recomputing all literals of \emph{K$_{i}$} while calculating \emph{U$_{i}$}, we
can use them to speed up the process. Note that the actual least fixpoint of \emph{U$_{i}$}
is the union of sets \emph{K$_{i}$} and \emph{opt\_lfp(P, K$_{i}$, K$_{i}$)}, as the optimized least
fixpoint computes only new literals (which are not included in given \emph{I}).

Since \emph{K$_{i-1}$} $\subseteq$ \emph{K$_{i}$}, for $i\geq1$ (see Lemma~\ref{Monotonicity}),
the computation of \emph{K$_{i}$} can start from \emph{K$_{i-1}$}, namely \emph{I} $=$
\emph{K$_{i-1}$}. Once \emph{opt\_lfp(P, K$_{i-1}$, U$_{i-1}$)} is computed, we append it
to our previously stored knowledge \emph{K$_{i-1}$}, resulting in \emph{K$_{i}$}. In addition,
a WFS fixpoint is reached when \emph{K$_{i-1}$} $=$ \emph{K$_{i}$}, namely when \emph{K$_{i-1}$} and
\emph{K$_{i}$} have the same number of literals.
\begin{proof}\label{def:WFS_fixpoint_condition}
If K$_{i-1}$ $=$ K$_{i}$, for $i\geq1$, then \\
U$_{i-1}$ $=$ K$_{i-1}$ $\cup$ opt\_lfp(P, K$_{i-1}$, K$_{i-1}$)  $=$ K$_{i}$ $\cup$ opt\_lfp(P, K$_{i}$, K$_{i}$) $=$ U$_{i}$ \\
Thus, fixpoint is reached as (K$_{i-1}$,U$_{i-1}$) =(K$_{i}$,U$_{i}$).
\end{proof}

According to Theorem~\ref{Correctness_of_AFP}, having reached WFS fixpoint at step $i$,
we can determine which literals are true, undefined and false as follows: (a) {\bf true}
literals, denoted by K$_{i}$, (b) {\bf undefined} literals, denoted by
U$_{i}$ $-$ K$_{i}$ and (c) {\bf false} literals, denoted by BASE(P) $-$ U$_{i}$.

Although for \emph{K$_{i}$} calculation only new literals are inferred during each ``inference step'',
for \emph{U$_{i}$} we have to recalculate a subset of literals that can be found in \emph{U$_{i-1}$}, as
literals in \emph{U$_{i-1}$} $-$ \emph{K$_{i-1}$} are discarded prior to the computation of \emph{U$_{i}$}.
However, the computational overhead coming from the calculation of \emph{opt\_lfp(P, K$_{i}$,
K$_{i}$)} reduces over time since the set of literals in \emph{U$_{i}$} $-$ \emph{K$_{i}$} becomes
smaller after each ``inference step'' due to \emph{K$_{i-1}$} $\subseteq$ \emph{K$_{i}$} and \emph{U$_{i-1}$}
$\supseteq$ \emph{U$_{i}$}, for $i\geq1$, (see Lemma~\ref{Monotonicity}).

We may further optimize our approach by minimizing the amount of stored literals.
A naive implementation would require the storage of up to four overlapping sets of literals (K$_{i-1}$, U$_{i-1}$, K$_{i}$,
U$_{i}$). However, as \emph{K$_{i}$} $\subseteq$ \emph{U$_{i}$}, while calculating \emph{U$_{i}$},
we need to store in our knowledge base only the sets \emph{K$_{i}$} and \emph{opt\_lfp(P, K$_{i}$,
K$_{i}$)}, since \emph{U$_{i}$}~$=$~\emph{K$_{i}$}~$\cup$~\emph{opt\_lfp(P, K$_{i}$, K$_{i}$)}.

As \emph{K$_{i-1}$} $\subseteq$ \emph{K$_{i}$}, for the calculation of \emph{K$_{i}$}, we need to store in
our knowledge base only three sets of literals, namely: (a) \emph{K$_{i-1}$},
(b) \emph{U$_{i-1}$} $-$ \emph{K$_{i-1}$} $=$ \emph{opt\_lfp(P, K$_{i-1}$, K$_{i-1}$)} and (c) currently
calculating least fixpoint \emph{opt\_lfp(P, K$_{i-1}$, U$_{i-1}$)}. All newly inferred literals
in \emph{opt\_lfp(P, K$_{i-1}$, U$_{i-1}$)}, are added to \emph{K$_{i}$} (replacing our prior knowledge
about \emph{K$_{i-1}$}), while literals in \emph{U$_{i-1}$} - \emph{K$_{i-1}$} $=$ \emph{opt\_lfp(P, K$_{i-1}$, K$_{i-1}$)}
are deleted, if fixpoint is not reached, as they cannot be used for the computation of \emph{U$_{i}$}.

A WFS fixpoint is reached when \emph{K$_{i-1}$} = \emph{K$_{i}$}, namely when no new literals
are derived during the calculation of \emph{K$_{i}$}, which practically is the calculation
of \emph{opt\_lfp(P,~K$_{i-1}$,~U$_{i-1}$)}. Since (K$_{i-1}$,U$_{i-1}$) $=$ (K$_{i}$,U$_{i}$),
we return the sets of literals \emph{K$_{i-1}$} and \emph{U$_{i-1}$}, representing our
fixpoint knowledge base.

In practice, the maximum amount of stored data occurs while calculating \emph{K$_{i}$}, for $i\geq1$,
where we need to store three sets of literals, namely: (a) \emph{K$_{i-1}$}, (b) \emph{U$_{i-1}$} $-$
\emph{K$_{i-1}$} and (c) \emph{opt\_lfp(P, K$_{i-1}$, U$_{i-1}$)}, requiring significantly less
storage space compared to the naive implementation.

\subsection{Computational Impact of Safety}\label{sec:Final_remarks}

In this paper, we follow the alternating fixpoint procedure, over safe WFS programs, in order to avoid full materialization
of or reasoning over the Herbrand base for any predicate. Storing or performing reasoning over the entire
Herbrand base may easily become prohibiting even for small datasets, and thus, not applicable to big data.

Apart from the semantic motivation of the safety requirement outlined in Section~\ref{sec:Well-Founded_Semantics}, it also has considerable impact on the computational method followed in this paper.
Recall that safety requires that
each variable in a rule must occur (also) in a positive subgoal. If this
safety condition is not met, an anti-join is no longer a single lookup between
the positive goal and a negative subgoal, but a comparison between a subset of the
Herbrand base and a given set of literals \emph{J}. An efficient implementation for such
computation is yet to be defined and problematic, as illustrated next.

Consider the following program:
\begin{center}
	\begin{tabular}{ l }
        p(X,Y) $\leftarrow$ a(X,Y), {\bf not} b(Y,Z). \\
        q(X,Y) $\leftarrow$ c(X,U), {\bf not} d(W,U), {\bf not} e(U,Y).
	\end{tabular}
\end{center}
For the first rule, each (\emph{X,Y}) in \emph{a(X,Y)}
is included in the final goal (\emph{p(X,Y)}) only if for a given \emph{Y}, there
is a \emph{Z} the in Herbrand universe such that \emph{b(Y,Z)} does not belong to \emph{J}.
For the second rule, for each (\emph{X,Y}) that is included in the final goal (\emph{q(X,Y)})
there should be a literal \emph{c(X,U)} that does not match neither \emph{d(W,U)} on \emph{U},
for any \emph{W} in Herbrand universe, nor \emph{e(U,Y)} on \emph{U}, for any \emph{Y} in
Herbrand universe. Thus, we need to perform reasoning over a subset of the Herbrand
base for \emph{b(Y,Z)}, \emph{d(W,U)} and \emph{e(U,Y)} in order to find the nonmatching literals.

\section{Experimental results}\label{sec:Experimental_results}

%
%
%

{\bf Methodology.}
In order to evaluate our approach, we surveyed available benchmarks
in the literature. In~(\cite{Liang:2009:OAP:1526709.1526790}), the authors
evaluate the performance of several rule engines on data that fit in
main memory. However, our approach is targeted on data that exceed the
capacity of the main memory. Thus, we follow the proposed methodology
in~(\cite{Liang:2009:OAP:1526709.1526790}) while adjusting several parameters.
In~(\cite{Liang:2009:OAP:1526709.1526790}) \emph{loading} and
\emph{inference} time are separated, focusing on inference time. However, for our approach such
a separation is difficult as loading and inference time may overlap.

We evaluate our approach considering \emph{default negation} by applying the
\emph{win-not-win} test and merge \emph{large (anti-)join tests} with \emph{datalog recursion}
and \emph{default negation}, creating a new test called \emph{transitive
closure with negation}. Other metrics in~(\cite{Liang:2009:OAP:1526709.1526790}), such
as \emph{indexing}, are not supported by the MapReduce framework, while all
optimizations and cost-based analysis were performed manually.

{\bf Platform.}
We have implemented our experiments using the Hadoop MapReduce
framework\footnote{\url{http://hadoop.apache.org/mapreduce/}}, version 1.2.1.
We have performed experiments on a cluster of the University of
Huddersfield. The cluster consists of 8 nodes (one node was allocated as ``master''
node), using a Gigabit Ethernet interconnect. Each node was equipped
with 4~cores running at 2.5GHz, 8GB RAM and 250GB of storage space.

{\bf Evaluation tests.}
The \emph{win-not-win} test~(\cite{Liang:2009:OAP:1526709.1526790})
consists of a single rule, where \emph{move} is the base relation:
\begin{center}
	\begin{tabular}{ c }
        win(X) $\leftarrow$ move(X,Y), {\bf not} win(Y).
	\end{tabular}
\end{center}

We test the following data distributions:
\begin{itemize}
  \item the base facts form a cycle: \{move(1,2), ..., move(i, i+1), ..., move(n-1,n), move(n,1)\}.
  \item the data is tree-structured: \{move(i, 2*i), move(i, 2*i+1) $|$ 1 $\leq$ i $\leq$ n\}.
\end{itemize}

We used four cyclic datasets and four tree-structured datasets with 125M, 250M, 500M and 1000M facts.

The \emph{transitive closure with negation} test consists of the following rule set, where \emph{b} is the base relation:
\begin{center}
	\begin{tabular}{ l l }
        tc(X,Y) $\leftarrow$ par(X,Y). & par(X,Y) $\leftarrow$ b(X,Y), {\bf not} q(X,Y). \\
        tc(X,Y) $\leftarrow$ par(X,Z), tc(Z,Y). & par(X,Y) $\leftarrow$ b(X,Y), b(Y,Z), {\bf not} q(Y,Z). \\
        q(X,Y) $\leftarrow$ b(Z,X), b(X,Y), {\bf not} q(Z,X).
	\end{tabular}
\end{center}

We test the following data distribution:
\begin{itemize}
  \item the base facts are chain-structured: \{b(i, i+k) $|$ 1 $\leq$ i $\leq$ n, k $<$ n\}. Intuitively,
  the $i$ values are distributed over $\lceil n/k \rceil$ levels, allowing $\lceil n/k \rceil~-~1$ joins in
  the formed chain.
\end{itemize}

The \emph{transitive closure with negation} test allows for comparing the performance of the naive and
the optimized WFS fixpoint calculation when the computation of lfp(\emph{T$_{P,J}$(I)})
starts from \emph{I} $= \emptyset$ and \emph{I} $\neq \emptyset$ respectively. For \emph{U$_{i}$} and
\emph{K$_{i+1}$}, for $i\geq0$, the optimized implementation speeds up
the process by using, as input, the previously computed transitive closure of \emph{K$_{i}$}, while
the naive implementation comes with the overhead of recomputing previously inferred literals.
Intuitively, this test allows the subsequent computation of transitive closure that becomes larger after
each ``inference step''.

We used four chain-structured datasets for increasing number of joins in the initially formed
chain ($\lceil n/k \rceil~-~1$) with n = 125M, and k = 41.7M, 25M, 13.9M and 7.36M, and four
chain-structured datasets for a constant number of joins in the initially formed chain ($\lceil n/k \rceil~-~1$)
with n = 62.5M, 125M, 250M and 500M, and k = 12.5M, 25M, 50M and 100M respectively.

{\bf Results.}
We can identify four main factors that affect the performance of our approach:
(a) \emph{number of facts}, affecting the input size, (b) \emph{number of rules},
affecting the output size, (c) \emph{data distribution}, affecting the number of
required MapReduce jobs, and (d) \emph{rule set structure}, affecting the number of
required MapReduce jobs.

\begin{figure}[t]
  \begin{minipage}[t]{0.48\textwidth}
    \centering
    \includegraphics[width=2.6in,height=1.3in]{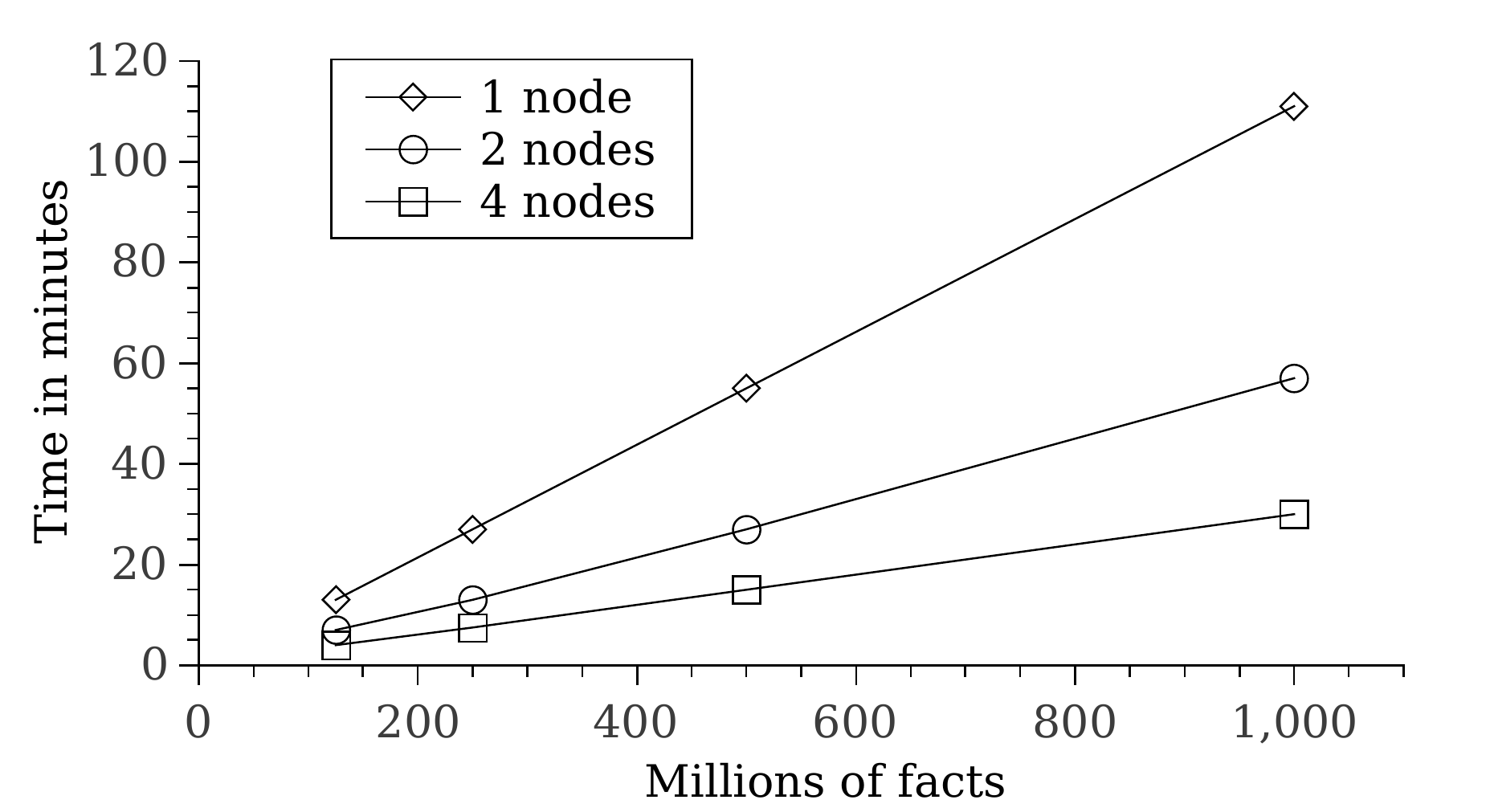}
   \caption{\emph{Win-not-win} test for cyclic datasets. Time in minutes as a function of dataset
   size, for various numbers of nodes.}
   \label{fig:win_not_win_cycle}
   \end{minipage}
  \hspace*{\fill} 
  \begin{minipage}[t]{0.48\textwidth}
    \centering
    \includegraphics[width=2.6in,height=1.3in]{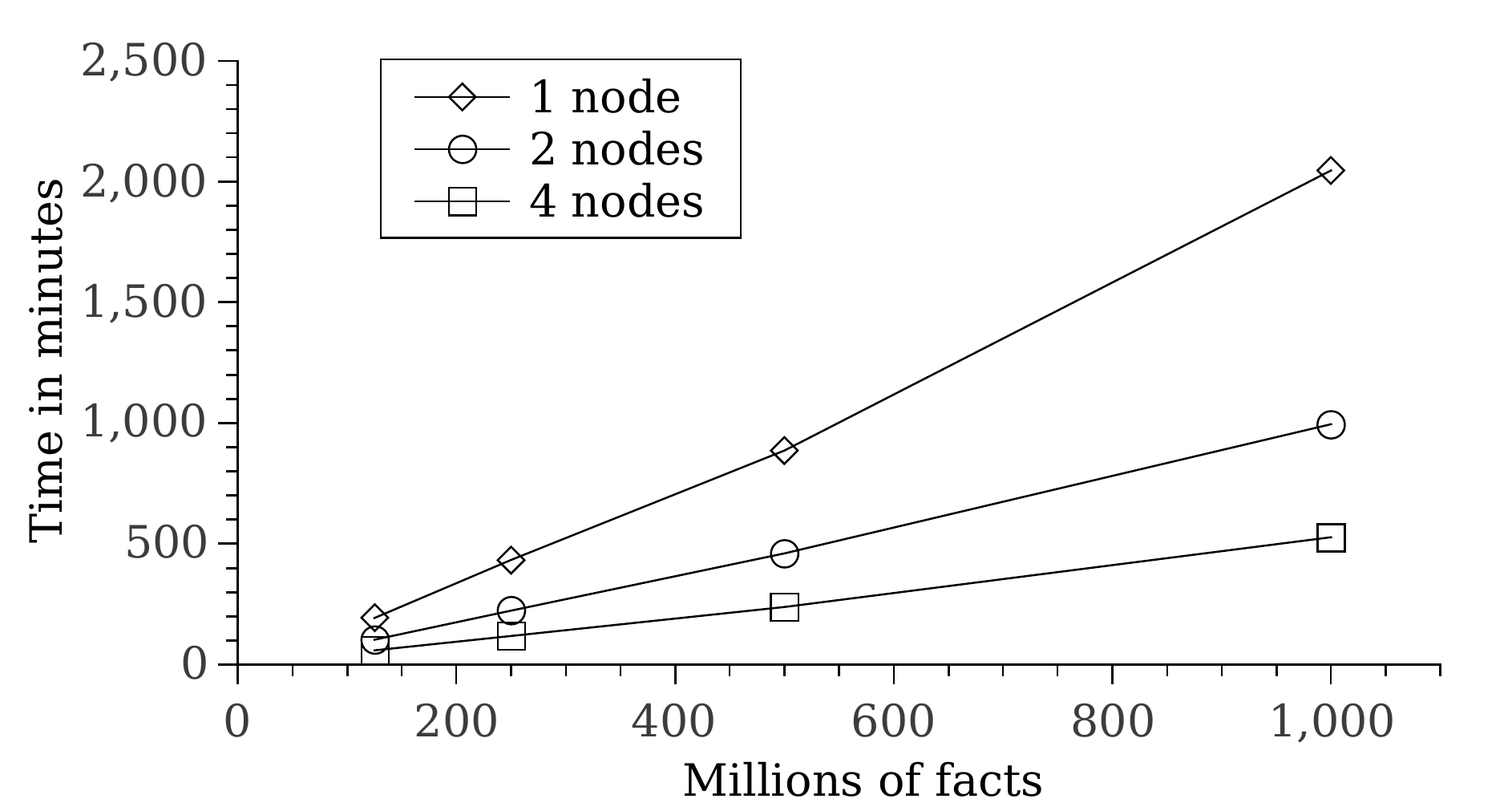}
    \caption{\emph{Win-not-win} test for tree-structured datasets. Time in minutes as a function of dataset
    size, for various numbers of nodes.}
    \label{fig:win_not_win_tree}
  \end{minipage}
\end{figure}

Figure~\ref{fig:win_not_win_cycle} presents the runtimes of our system for the \emph{win-not-win} test over
cyclic datasets with input sizes up to 1~billion facts. In this case, our system scales linearly
with respect to both dataset size and number of nodes. This is attributed to the fact that the runtime per
MapReduce job scales linearly for increasing data sizes, while the number of jobs remains constant.

Figure~\ref{fig:win_not_win_tree} shows the runtimes of our system for the \emph{win-not-win} test over
tree-structured datasets with input sizes up to 1~billion facts. Our approach scales linearly
for increasing data sizes and number of nodes.

\begin{figure}[t]
  \begin{minipage}[t]{0.48\textwidth}
    \centering
    \includegraphics[width=2.6in,height=1.3in]{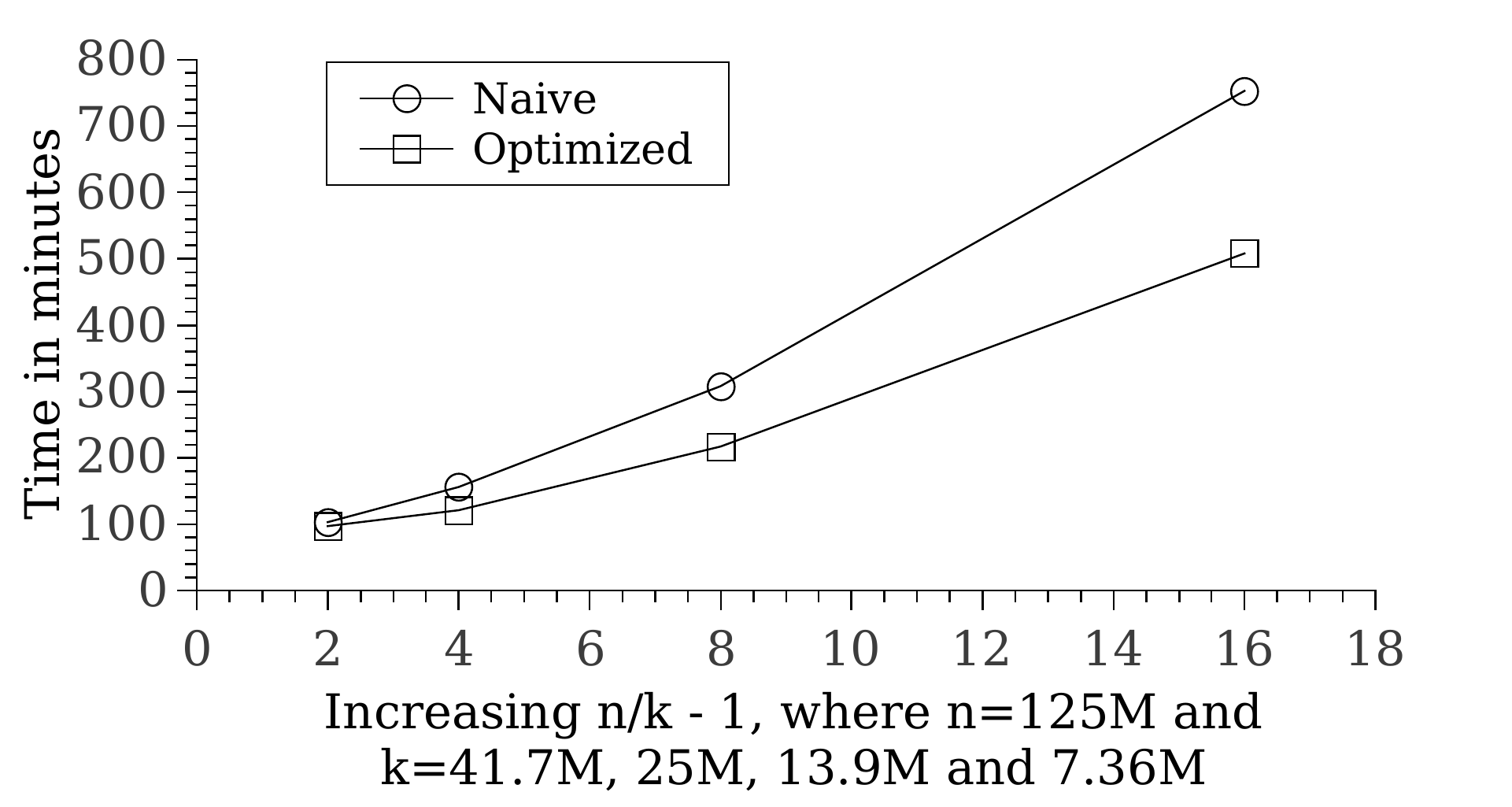}
    \caption{\emph{Transitive closure with negation} test for chain-structured datasets. Time in minutes
    for increasing $\lceil n/k \rceil~-~1$, 
    comparing naive and optimized WFS
    fixpoint calculation.}
    \label{fig:large_tc_with_neg_raising_levels}
   \end{minipage}
  \hspace*{\fill} 
  \begin{minipage}[t]{0.48\textwidth}
    \centering
    \includegraphics[width=2.6in,height=1.3in]{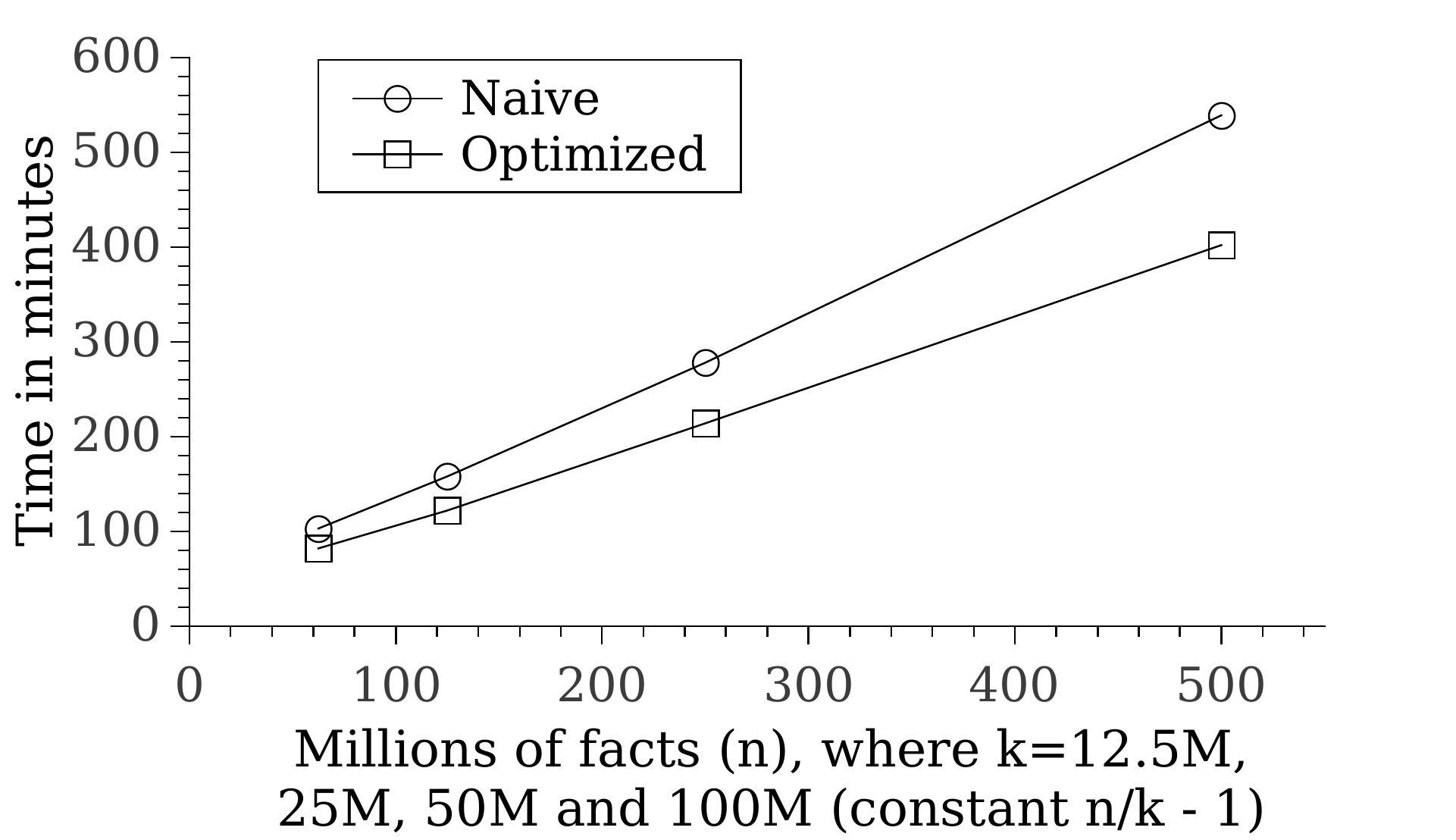}
    \caption{\emph{Transitive closure with negation} test for chain-structured datasets.
    Time in minutes 
    for constant $\lceil n/k \rceil~-~1$, comparing naive and optimized WFS
    fixpoint calculation.}
    \label{fig:large_tc_with_neg_stable_levels}
  \end{minipage}
\end{figure}

Figure~\ref{fig:large_tc_with_neg_raising_levels} depicts the scaling properties of our system for the \emph{transitive
closure with negation} test over chain-structured datasets, when run on 7~nodes.
Practically, transitive closure depends on the number of joins in the initially formed chain, which are equal
to $\lceil n/k \rceil~-~1$, namely 2, 4, 8 and 16, and thus, appropriate for scalability evaluation. The length
of the chain affects both the size of the transitive closure and the number of ``inference steps'', leading to
polynomial complexity. Note that our results are in line with Theorem~\ref{WFS_complexity}. Finally, the speedup of the optimized
over the naive implementation is higher for longer chains, since the naive implementation has to recompute larger
transitive closures.

Figure~\ref{fig:large_tc_with_neg_stable_levels} illustrates the scalability properties of our
system for the \emph{transitive closure with negation} test over chain-structured datasets for
constant number of joins in the initially formed chain, when run on 7~nodes. Our approach scales
linearly, both for naive and optimized implementation as the number of jobs remains constant,
while the runtime per job scales linearly for increasing number of facts.

\section{Conclusion and Future Work}\label{sec:Conclusion_and_Future_Work}
In this paper, we studied the feasibility of computing the well-founded semantics, while allowing
recursion through negation, over large amounts
of data. In particular, we proposed a parallel approach based on the MapReduce framework,
ran experiments for various rule sets and data sizes, and showed the performance speedup
coming from the optimized implementation when compared to a naive implementation. Our experimental
results indicate that this method can be applied to billions of facts.

In future work, we plan to study more complex knowledge representation
methods including Answer-Set programming~(\cite{Michael2008285}), and
RDF/S ontology evolution~(\cite{DBLP:conf/ecai/KonstantinidisFAC08})
and repair~(\cite{rousakis}). We believe that these complex forms of
reasoning do not fall under the category of ``embarrassingly
parallel'' problems for which MapReduce is designed, and thus, a more
complex computational model is required. Parallelization techniques
such as OpenMP\footnote{\url{http://openmp.org/wp/}} and Message
Passing Interface (MPI) may provide higher degree of flexibility than
the MapReduce framework, giving the opportunity to overcome arising
limitations. In fact, in Answer-Set programming, the system claspar
(\cite{DBLP:conf/lpnmr/GebserKKSS11}) uses MPI, but it needs a
preliminary grounding step, as it accepts only ground or propositional
programs. (\cite{DBLP:journals/tplp/PerriRS13}) uses POSIX threads on
shared memory for parallelized grounding. Combining these two
approaches and making them more data-driven would be an interesting
challenge.

\appendix

\section{MapReduce algorithms}\label{sec:MapReduce_algorithms}

In the appendix, we include the algorithms that are used in the running
examples of this paper. More specifically, Algorithm~\ref{wordcount} refers
to the wordcount example in Section~\ref{sec:MapReduce_Framework}.

\begin{algorithm}[htbp]
\begin{algorithmic}[1]
\Statex map(Long \emph{key}, String \emph{value}): \Comment \emph{key}: position in document
\ForAll {word \emph{w} $\in$ \emph{value}} \Comment \emph{value}: document line
\State emit(\emph{w}, ``1'');
\EndFor
\Statex
\Statex reduce(String \emph{key}, Iterator \emph{values}): \Comment \emph{key}: a word
\State int \emph{count} = 0; \Comment \emph{values}: list of counts
\ForAll {\emph{value} $\in$ \emph{values}}
\State \emph{count} += parseInt(\emph{value});
\EndFor
\State emit(\emph{key}, \emph{count});
\end{algorithmic}
\caption{Wordcount example}
\label{wordcount}
\end{algorithm}


In Section~\ref{sec:Positive_goal_calculation} we described the calculation
of the positive goal by applying a single join following Algorithm~\ref{Single_join}.

\begin{algorithm}[htbp]
\begin{algorithmic}[1]
\Statex map(Long \emph{key}, String \emph{value}):  \Comment \emph{key}: position in document (irrelevant)
\If {\emph{value.predicate} == ``a''}               \Comment \emph{value}: document line (literal in \emph{I})
\State     emit(\emph{value.Z},\{\emph{value.predicate},\emph{value.X}\});
\ElsIf {\emph{value.predicate} == ``b''}
\State     emit(\emph{value.Z},\{\emph{value.predicate},\emph{value.Y}\});
\EndIf
\Statex
\Statex reduce(String \emph{key}, Iterator \emph{values}): \Comment \emph{key}: matching argument
\State List \emph{a\_List} = $\emptyset$, \emph{b\_List} = $\emptyset$; \Comment \emph{values}: literals in \emph{I} for matching
\ForAll {\emph{value} $\in$ \emph{values}}
\If {\emph{value.predicate} == ``a''}
\State     \emph{a\_List}.add(\emph{value.X});
\ElsIf {\emph{value.predicate} == ``b''}
\State     \emph{b\_List}.add(\emph{value.Y});
\EndIf
\EndFor
\ForAll {\emph{a} $\in$ \emph{a\_List}}
\ForAll {\emph{b} $\in$ \emph{b\_List}}
\State emit(``ab(\emph{a.X},\emph{key.Z},\emph{b.Y})'',``'');
\EndFor
\EndFor
\end{algorithmic}
\caption{Single join}
\label{Single_join}
\end{algorithm}


Although we mentioned, in Section~\ref{sec:Positive_goal_calculation}, that
duplicate elimination should take place as soon as
possible in order to minimize overhead, the description of the algorithm was
deferred to this appendix. Duplicate elimination can be performed as described
in Algorithm~\ref{alg:Duplicate_elimination}. Practically, the $Map$ function
emits every inferred literal as the key, with an empty value. The MapReduce
framework performs grouping/sorting resulting in one group (of duplicates) for
each unique literal. Each group of duplicates consists of the unique literal
as the key and a set of empty values (with values being eventually ignored).
The actual duplicate elimination takes place during the reduce phase since for
each group of duplicates, we emit the (unique) inferred literal once, using the
key, while ignoring the values.

\begin{algorithm}[htbp]
\begin{algorithmic}[1]
\Statex map(Long \emph{key}, String \emph{value}): \Comment \emph{key}: position in document (irrelevant)
\State emit(\emph{value}, ``''); \Comment \emph{value}: document line (inferred literal)
\Statex
\Statex reduce(String \emph{key}, Iterator \emph{values}): \Comment \emph{key}: inferred literal
\State emit(\emph{key}, ``''); \Comment \emph{values}: empty values (not used)
\end{algorithmic}
\caption{Duplicate elimination}
\label{alg:Duplicate_elimination}
\end{algorithm}


Finally, the calculation of the final goal as described in Section~\ref{sec:Final_goal_calculation}
follows Algorithm~\ref{Anti_join}.

\begin{algorithm}[htbp]
\begin{algorithmic}[1]
\Statex map(Long \emph{key}, String \emph{value}):  \Comment \emph{key}: position in document (irrelevant)
\If {\emph{value.predicate} == ``ab''}               \Comment \emph{value}: document line (literal)
\State     emit(\{\emph{value.X},\emph{value.Z}\},\{\emph{value.predicate},\emph{value.Y}\});
\ElsIf {\emph{value.predicate} == ``c''}
\State     emit(\{\emph{value.X},\emph{value.Z}\},\emph{value.predicate});
\EndIf
\Statex
\Statex reduce(String \emph{key}, Iterator \emph{values}): \Comment \emph{key}: matching argument
\State List \emph{ab\_List} = $\emptyset$; \Comment \emph{values}: literals for matching
\ForAll {\emph{value} $\in$ \emph{values}}
\If {\emph{value.predicate} == ``ab''}
\State     \emph{ab\_List}.add(\emph{value.Y});
\ElsIf {\emph{value.predicate} == ``c''}
\State     \Return; \Comment matched by predicate \emph{c}
\EndIf
\EndFor
\ForAll {\emph{ab} $\in$ \emph{ab\_List}}
\State emit(``abc(\emph{key.X},\emph{key.Z},\emph{ab.Y})'',``'');
\EndFor
\end{algorithmic}
\caption{Anti-join}
\label{Anti_join}
\end{algorithm}


\bibliographystyle{acmtrans}
\bibliography{iclp2014_final}

\begin{thebibliography}{}

\bibitem[\protect\citeauthoryear{Abiteboul, Hull, and Vianu}{Abiteboul
  et~al\mbox{.}}{1995}]{DBLP:books/aw/AbiteboulHV95}
{\sc Abiteboul, S.}, {\sc Hull, R.}, {\sc and} {\sc Vianu, V.} 1995.
\newblock {\em Foundations of Databases}.
\newblock Addison-Wesley.

\bibitem[\protect\citeauthoryear{Afrati and Ullman}{Afrati and
  Ullman}{2010}]{DBLP:conf/edbt/AfratiU10}
{\sc Afrati, F.~N.} {\sc and} {\sc Ullman, J.~D.} 2010.
\newblock Optimizing joins in a map-reduce environment.
\newblock In {\em EDBT}, {I.~Manolescu}, {S.~Spaccapietra}, {J.~Teubner},
  {M.~Kitsuregawa}, {A.~L{\'e}ger}, {F.~Naumann}, {A.~Ailamaki}, {and}
  {F.~{\"O}zcan}, Eds. ACM International Conference Proceeding Series, vol.
  426. ACM, 99--110.

\bibitem[\protect\citeauthoryear{Brass, Dix, Freitag, and Zukowski}{Brass
  et~al\mbox{.}}{2001}]{bradixfrezuk99}
{\sc Brass, S.}, {\sc Dix, J.}, {\sc Freitag, B.}, {\sc and} {\sc Zukowski, U.}
  2001.
\newblock Transformation-based bottom-up computation of the well-founded model.
\newblock {\em Theory and Practice of Logic Programming\/}~{\em 1,\/}~5,
  497--538.

\bibitem[\protect\citeauthoryear{Cluet and Moerkotte}{Cluet and
  Moerkotte}{1994}]{Cluet94classificationand}
{\sc Cluet, S.} {\sc and} {\sc Moerkotte, G.} 1994.
\newblock Classification and optimization of nested queries in object bases.
\newblock Tech. rep.

\bibitem[\protect\citeauthoryear{Dean and Ghemawat}{Dean and
  Ghemawat}{2004}]{Dean:2004:MSD:1251254.1251264}
{\sc Dean, J.} {\sc and} {\sc Ghemawat, S.} 2004.
\newblock {M}ap{R}educe: simplified data processing on large clusters.
\newblock In {\em Proceedings of the 6th conference on Symposium on Opearting
  Systems Design \& Implementation - Volume 6}. USENIX Association, Berkeley,
  CA, USA, 10--10.

\bibitem[\protect\citeauthoryear{Fensel, van Harmelen, Andersson, Brennan,
  Cunningham, Valle, Fischer, Huang, Kiryakov, il~Lee, Schooler, Tresp, Wesner,
  Witbrock, and Zhong}{Fensel
  et~al\mbox{.}}{2008}]{DBLP:conf/semco/FenselHABCVFHKLSTWWZ08}
{\sc Fensel, D.}, {\sc van Harmelen, F.}, {\sc Andersson, B.}, {\sc Brennan,
  P.}, {\sc Cunningham, H.}, {\sc Valle, E.~D.}, {\sc Fischer, F.}, {\sc Huang,
  Z.}, {\sc Kiryakov, A.}, {\sc il~Lee, T.~K.}, {\sc Schooler, L.}, {\sc Tresp,
  V.}, {\sc Wesner, S.}, {\sc Witbrock, M.}, {\sc and} {\sc Zhong, N.} 2008.
\newblock Towards {L}ar{KC}: {A} {P}latform for {W}eb-{S}cale {R}easoning.
\newblock In {\em ICSC}. 524--529.

\bibitem[\protect\citeauthoryear{Gebser, Kaminski, Kaufmann, Schaub, and
  Schnor}{Gebser et~al\mbox{.}}{2011}]{DBLP:conf/lpnmr/GebserKKSS11}
{\sc Gebser, M.}, {\sc Kaminski, R.}, {\sc Kaufmann, B.}, {\sc Schaub, T.},
  {\sc and} {\sc Schnor, B.} 2011.
\newblock Cluster-based asp solving with {\it claspar}.
\newblock In {\em Logic Programming and Nonmonotonic Reasoning - 11th
  International Conference, LPNMR 2011, Vancouver, Canada, May 16-19, 2011.
  Proceedings}, {J.~P. Delgrande} {and} {W.~Faber}, Eds. Lecture Notes in
  Computer Science, vol. 6645. 364--369.

\bibitem[\protect\citeauthoryear{Gelder, Ross, and Schlipf}{Gelder
  et~al\mbox{.}}{1991}]{DBLP:journals/jacm/GelderRS91}
{\sc Gelder, A.~V.}, {\sc Ross, K.~A.}, {\sc and} {\sc Schlipf, J.~S.} 1991.
\newblock The well-founded semantics for general logic programs.
\newblock {\em J. ACM\/}~{\em 38,\/}~3, 620--650.

\bibitem[\protect\citeauthoryear{Gelfond}{Gelfond}{2008}]{Michael2008285}
{\sc Gelfond, M.} 2008.
\newblock Chapter 7 answer sets.
\newblock In {\em {H}andbook of {K}nowledge {R}epresentation}, {V.~L. F.~van
  Harmelen} {and} {B.~Porter}, Eds. Foundations of Artificial Intelligence,
  vol.~3. Elsevier, 285--316.

\bibitem[\protect\citeauthoryear{Goodman, Jimenez, Mizell, Al-Saffar, Adolf,
  and Haglin}{Goodman et~al\mbox{.}}{2011}]{DBLP:conf/esws/GoodmanJMAAH11}
{\sc Goodman, E.~L.}, {\sc Jimenez, E.}, {\sc Mizell, D.}, {\sc Al-Saffar, S.},
  {\sc Adolf, B.}, {\sc and} {\sc Haglin, D.~J.} 2011.
\newblock High-performance computing applied to semantic databases.
\newblock In {\em ESWC (2)}, {G.~Antoniou}, {M.~Grobelnik}, {E.~P.~B. Simperl},
  {B.~Parsia}, {D.~Plexousakis}, {P.~D. Leenheer}, {and} {J.~Z. Pan}, Eds.
  Lecture Notes in Computer Science, vol. 6644. Springer, 31--45.

\bibitem[\protect\citeauthoryear{Konstantinidis, Flouris, Antoniou, and
  Christophides}{Konstantinidis
  et~al\mbox{.}}{2008}]{DBLP:conf/ecai/KonstantinidisFAC08}
{\sc Konstantinidis, G.}, {\sc Flouris, G.}, {\sc Antoniou, G.}, {\sc and} {\sc
  Christophides, V.} 2008.
\newblock {A} {F}ormal {A}pproach for {RDF/S} {O}ntology {E}volution.
\newblock In {\em ECAI}, {M.~Ghallab}, {C.~D. Spyropoulos}, {N.~Fakotakis},
  {and} {N.~M. Avouris}, Eds. Frontiers in Artificial Intelligence and
  Applications, vol. 178. IOS Press, 70--74.

\bibitem[\protect\citeauthoryear{Kotoulas, Oren, and van Harmelen}{Kotoulas
  et~al\mbox{.}}{2010}]{DBLP:conf/www/KotoulasOH10}
{\sc Kotoulas, S.}, {\sc Oren, E.}, {\sc and} {\sc van Harmelen, F.} 2010.
\newblock {M}ind the data skew: distributed inferencing by speeddating in
  elastic regions.
\newblock In {\em WWW}, {M.~Rappa}, {P.~Jones}, {J.~Freire}, {and}
  {S.~Chakrabarti}, Eds. ACM, 531--540.

\bibitem[\protect\citeauthoryear{Liang, Fodor, Wan, and Kifer}{Liang
  et~al\mbox{.}}{2009}]{Liang:2009:OAP:1526709.1526790}
{\sc Liang, S.}, {\sc Fodor, P.}, {\sc Wan, H.}, {\sc and} {\sc Kifer, M.}
  2009.
\newblock Openrulebench: an analysis of the performance of rule engines.
\newblock In {\em Proceedings of the 18th international conference on World
  wide web}. WWW '09. ACM, New York, NY, USA, 601--610.

\bibitem[\protect\citeauthoryear{Liu, Qi, Wang, and Yu}{Liu
  et~al\mbox{.}}{2011}]{Liu:2011:LSF:2063016.2063043}
{\sc Liu, C.}, {\sc Qi, G.}, {\sc Wang, H.}, {\sc and} {\sc Yu, Y.} 2011.
\newblock {L}arge scale fuzzy p{D}* reasoning using mapreduce.
\newblock In {\em Proceedings of the 10th international conference on The
  semantic web - Volume Part I}. ISWC'11. Springer-Verlag, Berlin, Heidelberg,
  405--420.

\bibitem[\protect\citeauthoryear{Liu, Qi, Wang, and Yu}{Liu
  et~al\mbox{.}}{2012}]{DBLP:journals/cim/LiuQWY12}
{\sc Liu, C.}, {\sc Qi, G.}, {\sc Wang, H.}, {\sc and} {\sc Yu, Y.} 2012.
\newblock {R}easoning with {L}arge {S}cale {O}ntologies in fuzzy p{D}* {U}sing
  {M}ap{R}educe.
\newblock {\em IEEE Comp. Int. Mag.\/}~{\em 7,\/}~2, 54--66.

\bibitem[\protect\citeauthoryear{Mutharaju, Maier, and Hitzler}{Mutharaju
  et~al\mbox{.}}{2010}]{DBLP:conf/dlog/MutharajuMH10}
{\sc Mutharaju, R.}, {\sc Maier, F.}, {\sc and} {\sc Hitzler, P.} 2010.
\newblock {A} {M}ap{R}educe {A}lgorithm for {EL}+.
\newblock In {\em Description Logics}.

\bibitem[\protect\citeauthoryear{Nicolas}{Nicolas}{1982}]{DBLP:journals/acta/Nicolas82}
{\sc Nicolas, J.-M.} 1982.
\newblock Logic for improving integrity checking in relational data bases.
\newblock {\em Acta Informatica\/}~{\em 18}, 227--253.

\bibitem[\protect\citeauthoryear{Oren, Kotoulas, Anadiotis, Siebes, ten Teije,
  and van Harmelen}{Oren et~al\mbox{.}}{2009}]{marvin}
{\sc Oren, E.}, {\sc Kotoulas, S.}, {\sc Anadiotis, G.}, {\sc Siebes, R.}, {\sc
  ten Teije, A.}, {\sc and} {\sc van Harmelen, F.} 2009.
\newblock {M}arvin: {D}istributed reasoning over large-scale {S}emantic {W}eb
  data.
\newblock {\em J. Web Sem.\/}~{\em 7,\/}~4, 305--316.

\bibitem[\protect\citeauthoryear{Perri, Ricca, and Sirianni}{Perri
  et~al\mbox{.}}{2013}]{DBLP:journals/tplp/PerriRS13}
{\sc Perri, S.}, {\sc Ricca, F.}, {\sc and} {\sc Sirianni, M.} 2013.
\newblock Parallel instantiation of asp programs: techniques and experiments.
\newblock {\em Theory and Practice of Logic Programming\/}~{\em 13,\/}~2,
  253--278.

\bibitem[\protect\citeauthoryear{Roussakis, Flouris, and
  Christophides}{Roussakis et~al\mbox{.}}{2011}]{rousakis}
{\sc Roussakis, Y.}, {\sc Flouris, G.}, {\sc and} {\sc Christophides, V.} 2011.
\newblock {D}eclarative {R}epairing {P}olicies for {C}urated {KB}s.
\newblock In {\em HDMS}.

\bibitem[\protect\citeauthoryear{Soma and Prasanna}{Soma and
  Prasanna}{2008}]{DBLP:conf/icpp/SomaP08}
{\sc Soma, R.} {\sc and} {\sc Prasanna, V.~K.} 2008.
\newblock Parallel {I}nferencing for {OWL} {K}nowledge {B}ases.
\newblock In {\em ICPP}. IEEE Computer Society, 75--82.

\bibitem[\protect\citeauthoryear{Tachmazidis and Antoniou}{Tachmazidis and
  Antoniou}{2013}]{DBLP:conf/ruleml/TachmazidisA13}
{\sc Tachmazidis, I.} {\sc and} {\sc Antoniou, G.} 2013.
\newblock Computing the {S}tratified {S}emantics of {L}ogic {P}rograms over
  {B}ig {D}ata through {M}ass {P}arallelization.
\newblock In {\em RuleML}, {L.~Morgenstern}, {P.~S. Stefaneas}, {F.~L{\'e}vy},
  {A.~Wyner}, {and} {A.~Paschke}, Eds. Lecture Notes in Computer Science, vol.
  8035. Springer, 188--202.

\bibitem[\protect\citeauthoryear{Tachmazidis, Antoniou, Flouris, and
  Kotoulas}{Tachmazidis et~al\mbox{.}}{2012}]{DBLP:conf/kr/TachmazidisAFK12}
{\sc Tachmazidis, I.}, {\sc Antoniou, G.}, {\sc Flouris, G.}, {\sc and} {\sc
  Kotoulas, S.} 2012.
\newblock Towards {P}arallel {N}onmonotonic {R}easoning with {B}illions of
  {F}acts.
\newblock In {\em KR}, {G.~Brewka}, {T.~Eiter}, {and} {S.~A. McIlraith}, Eds.
  AAAI Press.

\bibitem[\protect\citeauthoryear{Tachmazidis, Antoniou, Flouris, Kotoulas, and
  McCluskey}{Tachmazidis
  et~al\mbox{.}}{2012}]{DBLP:conf/ecai/TachmazidisAFKM12}
{\sc Tachmazidis, I.}, {\sc Antoniou, G.}, {\sc Flouris, G.}, {\sc Kotoulas,
  S.}, {\sc and} {\sc McCluskey, L.} 2012.
\newblock Large-scale {P}arallel {S}tratified {D}efeasible {R}easoning.
\newblock In {\em ECAI}, {L.~D. Raedt}, {C.~Bessi{\`e}re}, {D.~Dubois},
  {P.~Doherty}, {P.~Frasconi}, {F.~Heintz}, {and} {P.~J.~F. Lucas}, Eds.
  Frontiers in Artificial Intelligence and Applications, vol. 242. IOS Press,
  738--743.

\bibitem[\protect\citeauthoryear{Urbani, Kotoulas, Massen, van Harmelen, and
  Bal}{Urbani et~al\mbox{.}}{2012}]{j.websem222}
{\sc Urbani, J.}, {\sc Kotoulas, S.}, {\sc Massen, J.}, {\sc van Harmelen, F.},
  {\sc and} {\sc Bal, H.} 2012.
\newblock Webpie: {A} {W}eb-scale parallel inference engine using
  {M}ap{R}educe.
\newblock {\em Web Semantics: Science, Services and Agents on the World Wide
  Web\/}~{\em 10,\/}~0.

\end{thebibliography}

\label{lastpage}
\end{document}